\documentclass[letterpaper, 10 pt, conference]{ieeeconf}  
\IEEEoverridecommandlockouts                              
\usepackage{graphics} 
\usepackage{epsfig} 
\usepackage{amsmath} 
\usepackage{amssymb}  
\usepackage{booktabs}
\usepackage{multirow} 
\usepackage{amsmath}
\usepackage{url} 
\usepackage{makecell}

\title{\LARGE \bf Hierarchical Image Matching for UAV Absolute Visual Localization via Semantic and Structural Constraints%
\author{Xiangkai Zhang$^{1,*}$, Xiang Zhou$^{2}$, Mao Chen$^{1}$, Yuchen Lu$^{1}$, Xu Yang$^{1}$, Zhiyong Liu$^{1}$
\thanks{$^{1}$ Institute of Automation, Chinese Academy of Sciences, Beijing, China. %
$^{2}$Guangxi University. %
$^{*}${\tt\small zhangxiangkai2023@ia.ac.cn}%
}}
}

\usepackage{eso-pic}
\AddToShipoutPicture*{
  \put(0, 770){%
    \parbox{\paperwidth}{%
      \centering
      \small This work has been submitted to the IEEE for possible publication. Copyright may be transferred without notice, \\
      after which this version may no longer be accessible.
    }
  }
}

\begin{document}

\maketitle
\thispagestyle{empty}
\pagestyle{empty}

\begin{abstract} Absolute localization, aiming to determine an agent's location with respect to a global reference, is crucial for unmanned aerial vehicles (UAVs) in various applications, but it becomes challenging when global navigation satellite system (GNSS) signals are unavailable. Vision-based absolute localization methods, which locate the current view of the UAV in a reference satellite map to estimate its position, have become popular in GNSS-denied scenarios. However, existing methods mostly rely on traditional and low-level image matching, suffering from difficulties due to significant differences introduced by cross-source discrepancies and temporal variations. To overcome these limitations, in this paper, we introduce a hierarchical cross-source image matching method designed for UAV absolute localization, which integrates a semantic-aware and structure-constrained coarse matching module with a lightweight fine-grained matching module. Specifically, in the coarse matching module, semantic features derived from a vision foundation model first establish region-level correspondences under semantic and structural constraints. Then, the fine-grained matching module is applied to extract fine features and establish pixel-level correspondences. Building upon this, a UAV absolute visual localization pipeline is constructed without any reliance on relative localization techniques, mainly by employing an image retrieval module before the proposed hierarchical image matching modules. Experimental evaluations on public benchmark datasets and a newly introduced CS-UAV dataset demonstrate superior accuracy and robustness of the proposed method under various challenging conditions, confirming its effectiveness. \end{abstract}
\section{Introduction}
In recent years, unmanned aerial vehicles (UAVs) have become indispensable across a variety of applications, such as military operations~\cite{military_uav}, transportation~\cite{transport_uav}, search-and-rescue missions~\cite{scherer2015autonomous, qi2016search}, etc. 
As UAVs are often deployed in unstructured and dynamic environments, ensuring robust localization is generally a necessary prerequisite for their task execution.

Absolute localization, aiming to determine an agent's position with respect to a fixed global reference rather than previous states, is crucial for unmanned aerial vehicles (UAVs) in various applications.
Although relative localization methods, such as inertial navigation systems (INS) and visual-inertial odometry (VIO), can provide localization results in the short term, they inevitably accumulate errors, resulting in significant drift during long-term missions~\cite{COUTURIER2021103666}. In contrast, absolute localization determines the UAV's position to a global reference, thereby correcting drift and ensuring long-term positional accuracy.

\begin{figure}[t] 
    \centering \includegraphics[width=\linewidth]{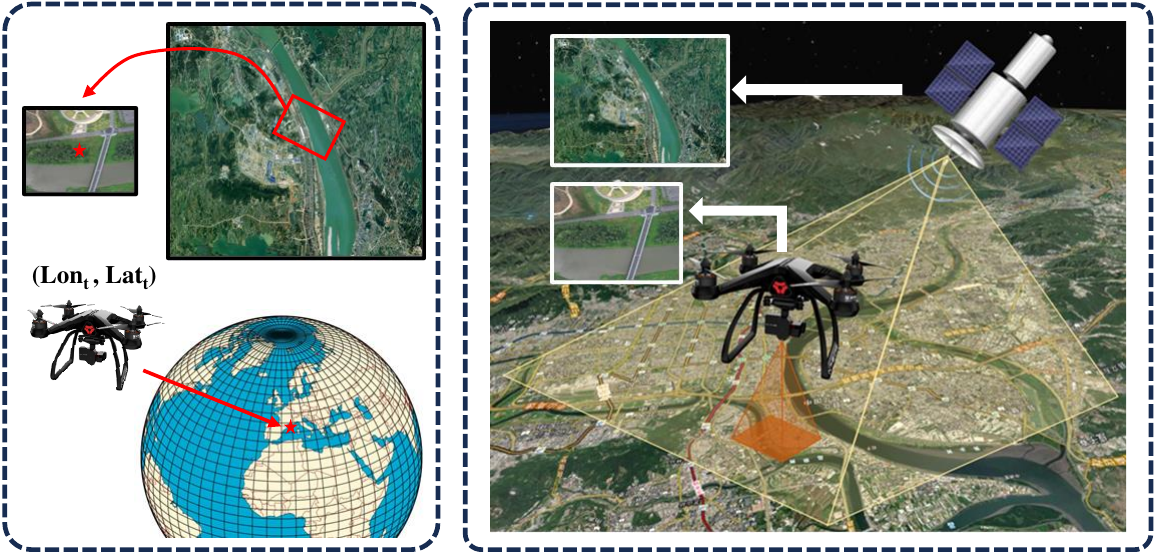} 
    \caption{An overview of the UAV absolute visual localization task, in which UAV-captured images are matched to a reference satellite map to estimate the UAV's position within a global coordinate system.} 
    \label{intro} 
\end{figure}
Existing UAV absolute localization techniques primarily rely on the global navigation satellite system (GNSS). Despite widespread adoption in daily applications, GNSS is vulnerable to signal interference and jamming, especially in challenging environments such as urban canyons, dense forests, or conflict zones, where physical obstructions and deliberate attacks may cause signal degradation or complete denial~\cite{UAV_survey}. Therefore, in such scenarios, an alternative absolute localization method without GNSS is necessary. 

Vision-based absolute localization methods, which locate the current view of the UAV in a reference satellite map to estimate its position (see Fig.~\ref{intro}), are considered promising alternatives to GNSS.
Image retrieval-based methods were first explored as an efficient way for UAV absolute visual localization~\cite{University-1652,Synthesis,duan2024scene,dai2023vision}, but they can only provide coarse localization results and need additional refinement. Matching-based methods estimate more precise UAV positions by directly matching UAV images and the whole reference satellite image to calculate their geometric correspondences~\cite{nassar2018deep,goforth2019gps,mughal2021assisting}. Although they perform well in small-scale flight scenarios, their effectiveness significantly diminishes as the UAV's operational area expands. 

Recently, the combined retrieval-and-matching methods have been proposed as a more reasonable design for UAV absolute visual localization, where image retrieval quickly selects candidate small-scale satellite image patches and image matching is subsequently applied to estimate precise positions~\cite{tingshua_uav,ye2024coarse,he2024aerialvl}. These methods require only a downward-facing camera and publicly available satellite imagery, offering a practical and cost-effective solution for UAV absolute localization.

\begin{figure*}[t]
  \centering
  \includegraphics[width=\textwidth]{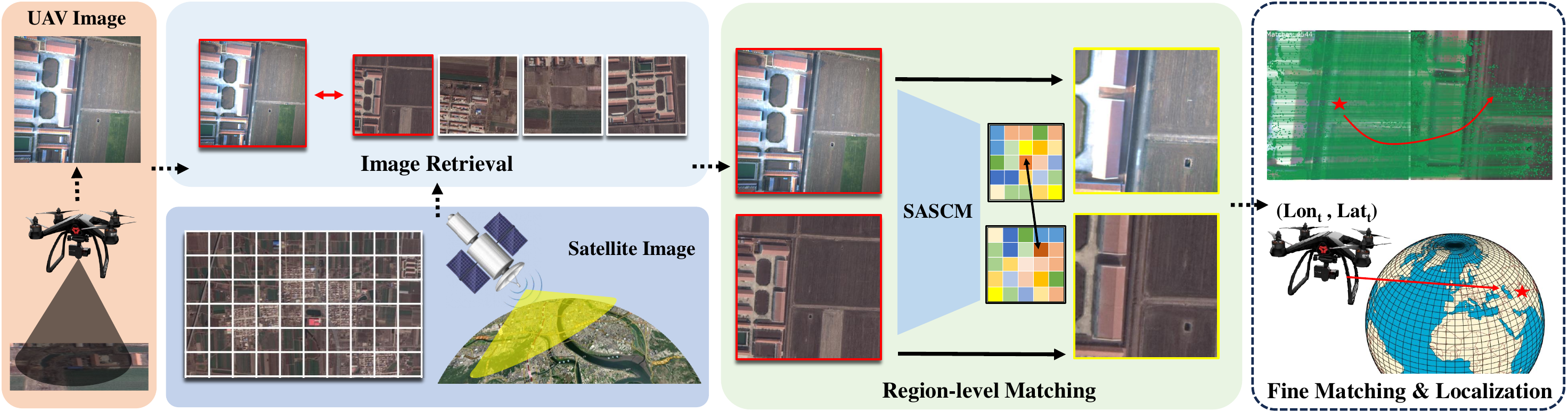}
  \caption{Overview of the proposed UAV absolute visual localization pipeline under GNSS-denied conditions. The pipeline consists of three key components: an image retrieval module that retrieves candidate satellite images based on visual similarity, a semantic-aware and structure-constrained matching module (SASCM) for identifying region-level matching, and a lightweight fine-grained matching module for fine matching and localization.}
  \label{overview}
\end{figure*}
However, despite improvements achieved by the combined retrieval-and-matching methods, matching UAV images with satellite images remains challenging in absolute visual localization. Existing methods mostly rely on traditional and low-level feature matching techniques, suffering from significant differences introduced by cross-source discrepancies and temporal variations.

To address these limitations, this work introduces a hierarchical cross-source image matching method to enhance UAV-satellite matching performance via semantic and structural constraints. 
Compared to low-level features, high-level semantic features offer a more abstract and invariant representation of scene content that remains relatively stable across different conditions~\cite{zhang2021semantic}. Leveraging semantic features can significantly enhance the robustness of UAV-to-satellite image matching. Yet, solely relying on semantic features may lead to ambiguous matches in scenes containing multiple regions with similar semantic content and lacking the fine-grained detail required for accurate pixel-level correspondence~\cite{Cadar2025}.
Therefore, our method integrates structural consistency constraints into semantic matching and adopts a hierarchical framework. Coarse semantic-level matching is first performed to mitigate the effects of cross-source discrepancies and temporal variations, while structural consistency is enforced to reduce matching ambiguity. To further improve matching accuracy by capturing detailed correspondences, the second stage utilizes a lightweight network to extract fine-grained features for pixel-level matching.

Based on this image matching method, a UAV absolute visual localization pipeline is constructed without any reliance on relative localization techniques, mainly by further employing an image retrieval module before the proposed hierarchical image matching modules, which consequently allows for robust UAV absolute visual localization in GNSS-denied scenarios. The primary contributions of this work are summarized as follows:

\begin{itemize} 
\item We propose a hierarchical cross-source image matching method that combines semantic-aware feature extraction with structure-constrained registration, significantly enhancing UAV-satellite image matching accuracy. 
\item Building upon the hierarchical matching method, we construct a UAV absolute visual localization pipeline without any reliance on relative localization techniques, mainly by employing an image retrieval module before the matching.
\item Comprehensive experimental evaluations on public benchmark datasets and a newly introduced CS-UAV dataset demonstrate the superior accuracy and robustness of the proposed method under various challenging conditions, confirming its effectiveness. 
\end{itemize}

\section{Related Work}
Methods for UAV absolute visual localization can be generally divided into three categories of retrieval-based methods, matching-based methods, and combined retrieval-and-matching methods.
\subsection{Retrieval-based methods}
Retrieval-based methods treat UAV visual localization as an image retrieval problem,  aiming to find the most similar georeferenced image (e.g., a satellite patch) to the UAV’s camera view. This can yield a coarse estimate of the UAV’s position by assigning it the georeferenced location of the retrieved image. 

As an efficient way for UAV absolute visual localization, various image-retrieval strategies have been explored. 
For instance, Shi et al .\cite{Synthesis} proposed a synthesis-based framework that generates satellite-style imagery from UAV views to address the domain gap between UAV and satellite views. Liu et al. \cite{duan2024scene} introduced a scene graph-based matching network tailored for UAV localization, which encodes spatial and semantic relationships within scenes. Dai et al. ~\cite{dai2023vision} further advanced this direction by incorporating metric learning into a vision-based self-positioning model for low-altitude urban UAV flights, enhancing feature discriminability and reducing modality discrepancies.
Despite their effectiveness, these methods inherently estimate the UAV’s position by identifying the most visually similar image from the database and assigning its georeferenced location as the UAV’s current position, which will inevitably introduce systematic errors.

\subsection{Matching-based methods}
Matching-based methods aim for more precise localization by establishing 2D-2D correspondences between the UAV image and a satellite image of the area. Instead of a coarse estimate, these methods directly match visual features between the UAV view and the reference map, then use geometric constraints (e.g., solving homography) to compute the UAV’s precise position. 

Nassar et al. \cite{nassar2018deep} introduced a CNN-based approach to improve UAV-satellite image registration by extracting robust feature representations and estimating geometric transformations.
Goforth et al. \cite{goforth2019gps} proposed a UAV localization framework that matches UAV-captured images with pre-existing satellite imagery to estimate position in GNSS-denied environments. By leveraging feature extraction and the Inverse Compositional Lucas-Kanade (ICLK) algorithm \cite{baker2004lucas}, their method achieves reliable geo-localization without external positioning signals.
Mughal et al. \cite{mughal2021assisting} proposed a method that leverages image contextual information to establish correspondences between UAV and satellite images. Similar to our work, their approach employs a neighborhood consistency network to capture structural information within image regions. 
Since these methods directly match UAV images with satellite reference images, they often struggle to achieve effective results when there is a significant difference in scale between the reference image and the UAV image.

\subsection{Combined retrieval-and-matching methods}
To integrate the advantages of retrieval-based and matching-based techniques, the combined methods were proposed as a more reasonable design for UAV absolute visual localization. In these methods, image retrieval quickly selects candidate small-scale satellite image patches and image matching is subsequently applied to estimate more precise positions.

Xu et al. \cite{tingshua_uav} introduced a robust end-to-end training framework that combines map retrieval and image matching techniques. Their approach designs a robust visual geo-localization pipeline that integrates a proposed deep learning-based imagery feature. The pipeline starts with image retrieval based on the deep feature encoding to initialize the localization process over large-scale maps. With the reuse of the same deep imagery feature, an image matching process enables real-time sequential localization. 
He et al. \cite{he2024aerialvl} proposed AerialVL by introducing a large-scale benchmark dataset for aerial-based visual localization and presented two combined methods serving as comparison baselines. 
Ye et al. \cite{ye2024coarse} proposed a coarse-to-fine geo-localization method for GNSS-denied UAVs using oblique-view images. They introduced Segments Direction Statistics (SDS) features for coarse bilateral retrieval, improved SuperPoint \cite{detone2018superpoint} with two additional modules for precise matching, and triangular pyramid constrained resection for accurate pixel-level localization.

Despite the significant improvements in adaptability and robustness achieved by the combined retrieval-and-matching methods, their image matching networks are often based on low-level image features and traditional frameworks. In real-world applications, these methods struggle with significant scene variations, leading to mismatches that compromise localization accuracy. To address this limitation, we redesign the matching process by decomposing it into a two-stage framework, leveraging the advanced semantic understanding capabilities of vision foundation models~\cite{oquab2023dinov2} and a hierarchical matching strategy, achieving more robust and accurate localization in complex real-world environments.
\begin{figure}[t]
  \centering
  \includegraphics[width=\linewidth]{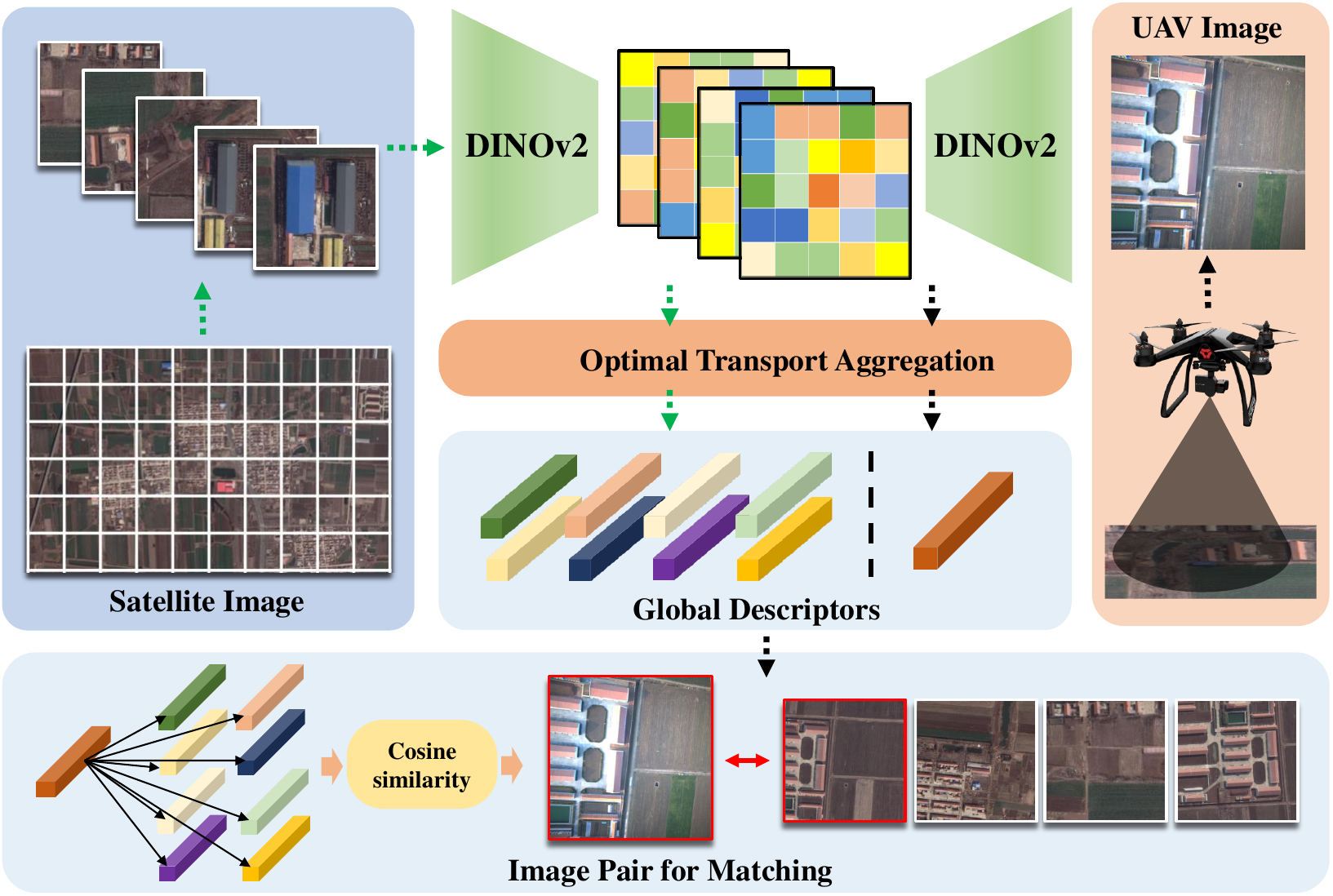}
  \caption{Diagram of the image retrieval module, where the green arrows indicate feature extraction and storage process in the offline stage, while the black arrows represent the retrieval process in the online stage.}
  \label{Image_Retrieval}
\end{figure}
\section{Proposed Approach}
We first introduce the main components of the proposed method, including an image retrieval module, a semantic-aware and structure-constrained matching module, and a lightweight fine-grained matching module. Then, we present the overall pipeline that integrates these modules into a unified absolute visual localization framework for UAVs, as illustrated in Fig.\ref{overview}.
\subsection{Image Retrieval Module}
We begin with a brief overview of the image retrieval module employed within our localization pipeline, which is designed to retrieve a satellite image covering the same geographic area as the image captured by the UAV. This ensures a high degree of overlap between the two images, providing a foundation for the subsequent matching process. 
Specifically, we adopt an off-the-shelf image retrieval model~\cite{salad} that is well-suited for UAV visual localization tasks. 

During the offline phase, the large-scale satellite image $I^S_{\text{ini}}$ is first cropped into a set of uniformly sized satellite images, denoted as $\mathcal{I}^S = \{ I^S_i, i=1,2,...,N\}$, which serves as a reference for the subsequent process. 
For each satellite image \( I^S_i \in \mathcal{I}^S\), we extract a feature map \( \mathbf{F}_i \) by a backbone network, and then apply optimal transport aggregation to obtain a compact and informative feature vector \( \mathbf{f}^S_i \), following the approach in~\cite{salad}.
Similarly, during the online phase, the image \( I^U_t \) captured by the UAV-mounted camera at time $t$ will undergo several preprocessing steps to ensure it aligns with the format of the satellite images in the database. This includes cropping and downsampling to maintain consistency in terms of resolution and size. Afterward, the feature extraction process is performed on the  \( I^U_t \), followed by the aggregation module to produce the corresponding feature vector \( \mathbf{f}^U_t \).
\begin{figure*}[t]
  \centering
  \includegraphics[width=\textwidth]{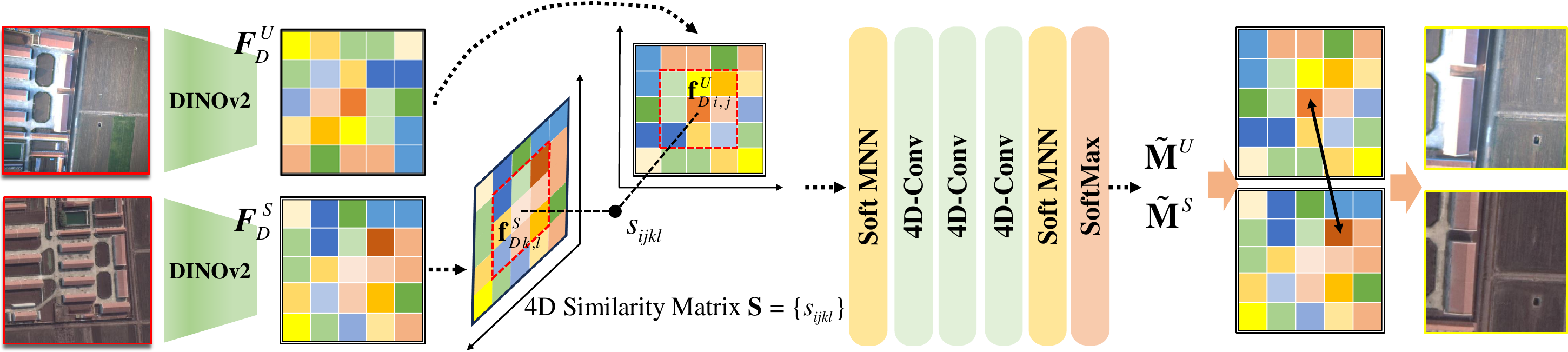}
  \caption{Architecture of the proposed Semantic-Aware and Structure-Constrained Matching Module (SASCM). This module first extracts dense semantic features using a vision foundation model (DINOv2), followed by the construction of a 4D correlation tensor encoding pairwise similarities across spatial positions. The Soft Mutual Nearest Neighbor (SoftMNN) filtering mechanism is applied to suppress ambiguous matches before and after the 4D convolutional layers, which capture local spatial consistency to identify reliable correspondences. A softmax operation is performed on \( \hat{\mathbf{M}}^U \) and \( \hat{\mathbf{M}}^S \) along the dimensions corresponding to images  \( I_{t}^U \)  and \( I_{ret}^S \), yielding the probabilistic representation of the matching matrix,}
  \label{sascm} 
\end{figure*}

Once the feature vector \( \mathbf{f}^U_t \) of the UAV image is obtained, we compute the similarity \( s_{t,i} \) between this vector and all feature vectors \( \mathbf{f}^S_i \) from the satellite images by
\begin{equation}
s_{t,i} = \frac{\mathbf{f}^U_t\mathbf{f}^S_i}{\|\mathbf{f}^U_t\|\|\mathbf{f}^S_i\|},
\end{equation}
where $\|\cdot\|$ denotes the Euclidean norm. After that, the satellite image with the highest similarity score is then retrieved as the reference image for matching:
\begin{equation}
ret = \arg\max_{i} s_{t,i} \ .
\end{equation}
The reference image \( I^S_{ret} \), together with the UAV image \( I^U_t \), is subsequently fed into the next stage of the visual localization pipeline. The entire image retrieval process is illustrated in Fig.\ref{Image_Retrieval}.
\subsection{Semantic-Aware and Structure-Constrained Matching Module (SASCM)}
\label{Method}
To mitigate the challenges posed by cross-source discrepancies and temporal variations, we propose a Semantic-Aware and Structure-Constrained Matching Module (SASCM). Instead of existing methods, which are sensitive to appearance changes, SASCM utilizes semantic features from a vision foundation model to establish robust region-level correspondences. In addition, structural consistency constraints are introduced to reduce ambiguities caused by repeated patterns and semantically similar regions. The detailed design of SASCM is presented below.

Given the UAV image \( I_{t}^U \) and the reference satellite image \( I_{ret}^S \) retrieved from the database as inputs, we first downsample both images to reduce the computational burden of subsequent feature extraction. The downsampled images are then passed through the DINOv2~\cite{oquab2023dinov2} to obtain dense feature maps \( \mathbf{F}^U_{\text{D}} = \{{\mathbf{f}^U_{\text{D}}}_{i,j}\} \in \mathbb{R}^{h \times w \times c} \) and \( \mathbf{F}^S_{\text{D}} = \{{\mathbf{f}^S_{\text{D}}}_{i,j}\} \in \mathbb{R}^{h \times w \times c} \).
We then compute the cosine similarity between each pair of feature vectors to form a 4D similarity matrix \( \mathbf{S} = \{s_{ijkl}\} \in \mathbb{R}^{h \times w \times h \times w} \):
\begin{equation}
s_{ijkl} = \frac{{\mathbf{f}^U_{\text{D}}}_{i,j} \cdot {\mathbf{f}^S_{\text{D}}}_{k,l}}{\|{\mathbf{f}^U_{\text{D}}}_{i,j}\| \cdot \|{\mathbf{f}^S_{\text{D}}}_{k,l}\|}.
\end{equation}
Then, a soft version of the mutual nearest neighbor filtering method (SoftMNN) as equation \eqref{softNN1} is employed to process the score matrix \( \mathbf{S} \) to get \( \mathbf{\hat{S}} \):
\begin{equation}\label{softNN1}
\hat{s}_{ijkl} = \frac{s_{ijkl}}{\text{max}_{ab}s_{abkl}} \cdot \frac{s_{ijkl}}{\text{max}_{cd}s_{ijcd}} \cdot s_{ijkl}.
\end{equation}
Within this constructed 4D matrix, an element \( \hat{s}_{ijkl} \) with indices \( i, j, k, l \) encodes the similarity between the positions \( (i,j) \) in \( \mathbf{F}^U \) and \( (k,l) \) in \( \mathbf{F}^S \) and elements in the vicinity of index \( i, j, k, l \) correspond to similarities between features in the local neighborhoods of descriptor \( {\mathbf{f}^U_{\text{D}}}_{i,j}\) and those in the local neighborhoods of descriptor \( {\mathbf{f}^S_{\text{D}}}_{k,l} \), as illustrated in Fig.\ref{sascm}. 

Although finding the correct match from the large number of elements in the matrix is a challenging task, based on the previously discussed neighborhood correspondence, we can reasonably infer that a correct match should have a set of supporting matches within its surrounding neighborhood in the 4D matrix. In contrast, an incorrect match would not exhibit this characteristic. In other words, by examining the features of a given point and its surrounding values in the 4D matrix, it is possible to filter and select the match. Specifically, we employ three convolutional layers, each consisting of 4D convolutional kernels, to process the score matrix, with the output of each layer passing through a ReLU activation function. This ultimately results in a 4D matching matrix, with the detailed network architecture shown in Fig.\ref{sascm}  and equation \eqref{N}-\eqref{M}:
\begin{equation}\label{N}
\mathcal{N}(\mathbf{\hat{S}}) = \left [\text{ReLU}(\text{4D-Conv}(\mathbf{\hat{S}})) \right ]_{\times 3},
\end{equation}
\begin{equation}\label{M}
\mathbf{M}=\mathcal{N}(\mathbf{\hat{S}})+\mathcal{N}(\mathbf{\hat{S}}^T).
\end{equation}

After obtaining the network output $ \mathbf{M} $, a SoftMNN is first applied to refine the output further while ensuring differentiability, resulting in $\hat{\mathbf{M}} = \text{SoftMNN}(\mathbf{M})$, where an element $\hat{m}_{ijkl}$ is defined as:
\begin{equation}\label{softNN}
\hat{m}_{ijkl} = \frac{m_{ijkl}}{\text{max}_{ab}m_{abkl}} \cdot \frac{m_{ijkl}}{\text{max}_{cd}m_{ijcd}} \cdot m_{ijkl}.
\end{equation}
Subsequently, a softmax operation as \eqref{softmax}-\eqref{softmax2} is performed on \( \hat{\mathbf{M}} \) along the dimensions corresponding to images  \( I_{t}^U \)  and \( I_{ret}^S \), yielding the probabilistic representation of the matching matrix, \( \widetilde{\mathbf{M}}^U \) and \( \widetilde{\mathbf{M}}^S \):
\begin{equation}\label{softmax}
{\tilde{m}}^U_{ijkl} = \frac{\text{exp} \\( \hat{m}_{ijkl}\\) } { \sum_{ab}\text{exp} \\( m_{abkl} \\)},
\end{equation}
\begin{equation}\label{softmax2}
{\tilde{m}}^S_{ijkl} = \frac{\text{exp} \\( \hat{m}_{ijkl}\\) } { \sum_{cd}\text{exp} \\( m_{ijcd} \\)}.
\end{equation}

In the inference phase, a hard-assignment constraint is applied to  \( \widetilde{\mathbf{M}}^U \) and \(  \widetilde{\mathbf{M}}^S\) to find the final point-to-point matching relationships.
During the training phase, a weakly supervised training approach consistent with that in \cite{rocco2020ncnet} is adopted, where the network is guided by the loss function defined as:
\begin{equation}\label{loss}
\mathcal{L}(I^U, I^S )=-y \left[  \text{Mean}\left( \widetilde{\mathbf{M}}^U \right)+ \text{Mean}\left( \widetilde{\mathbf{M}}^U \right) \right],
\end{equation}
in which the training pairs \( (I^U, I^S ) \) are positive pairs labeled with \(y = +1\), or negative pairs with \(y = -1\), and \(\text{Mean}(\cdot)\) denotes calculating the mean matching scores over all hard-assigned matches.

Although the point-level correspondences are obtained, the final matching results represent region-to-region correspondences rather than precise point-to-point matches. This is because the above process involves downsampling operations, and the feature maps extracted using the DINOv2 model have a resolution of only \( 1/14 \) of the input image.  

Based on the previous assumption that the UAV camera maintains a strictly downward-facing orientation, the central region of the UAV image can be regarded as corresponding to the actual geographic location of the UAV in terms of latitude and longitude. 
Therefore, after obtaining the above matching results, we focus on the correspondences of the center point of feature map \( \mathbf{F}^U_{\text{D}} \) and its surrounding points in \( \mathbf{F}^S_{\text{D}} \). 
These correspondences are then mapped back to the original image resolution, yielding region-level matching between the two input images, which can reduce the occurrence of mismatches and computational redundancy by concentrating only on the central region.

\subsection{Lightweight Fine-grained Matching Module}
The final stage of the matching process involves a fine-grained matching module that focuses on low-level texture features. This lightweight module performs precise detection and alignment between the keypoints in the central region  \( A \)  of the UAV image and the corresponding region  \(B\) in the satellite image, which are obtained from SASCM. By refining the initial coarse match, this step can improve the overall localization precision.

\begin{figure}[t]
      \centering
      \includegraphics[width=\linewidth]{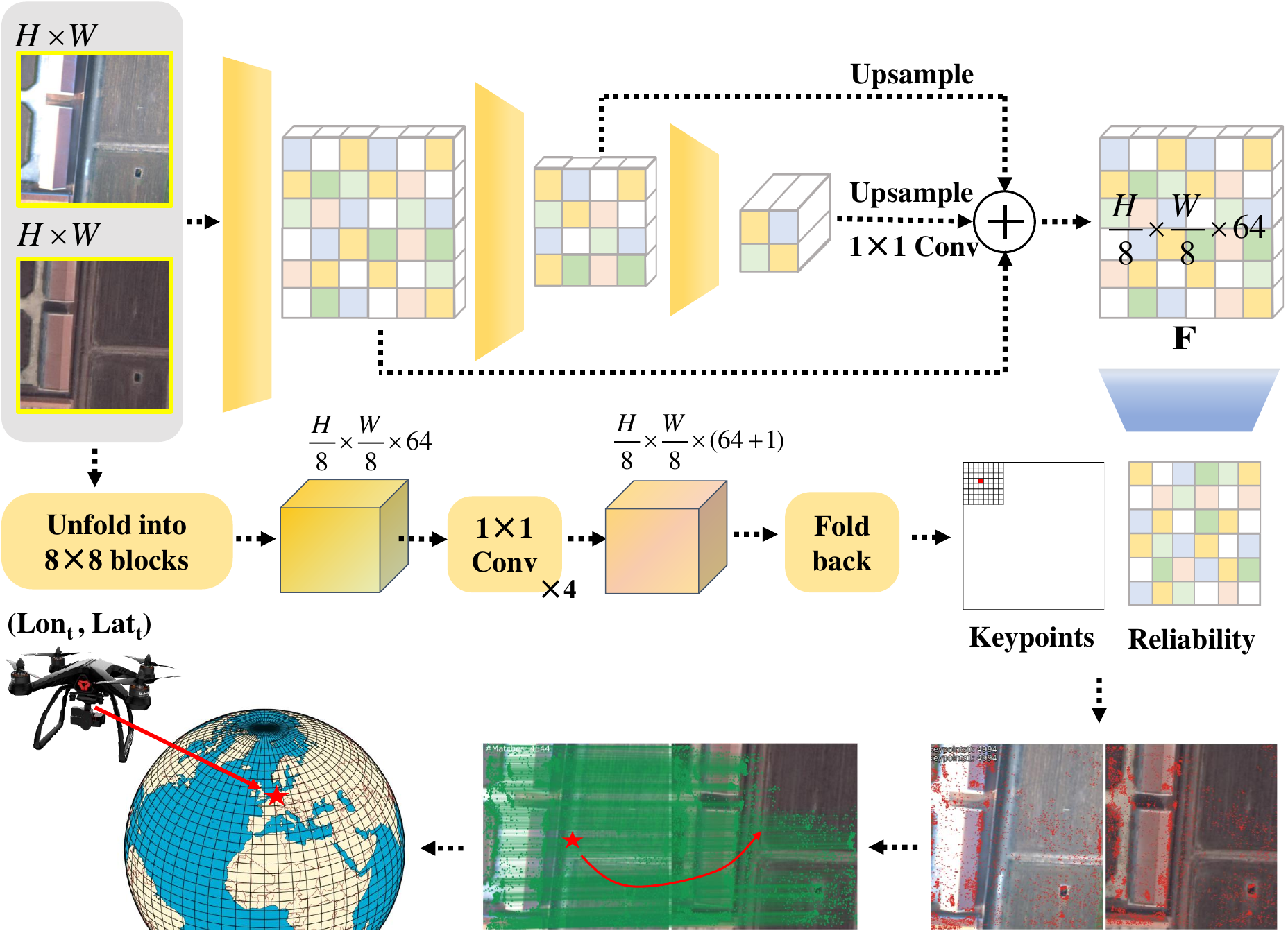}
      \caption{ Lightweight fine-grained matching module based on a convolutional keypoint detection and description network for precise texture-level alignment.}
      \label{Lightweight}
   \end{figure}

We adopt a network architecture similar to that described in \cite{potje2024xfeat}, as illustrated in Fig.\ref{Lightweight}. The feature extractor of the network consists of several conventional blocks, each containing multiple basic layers composed of 2D convolutional layer, ReLU, and Batch Normalization. By adjusting the kernel size and stride, the network extracts multi-level (\( H/8\times W/8, H/16\times W/16,H/32\times W/32\)) features from the input image area \( A \in \mathbb{R}^{H \times W \times 3} \). After that, a final feature fusion block is applied to generate the feature map \( \mathbf{F}_A \in \mathbb{R}^{H/8 \times W/8 \times 64}\). Based on this feature map, a keypoint descriptor \( \mathbf d_{ij} \) can be obtained by interpolation. 

Then, \(\mathbf{F}_A \) will be subsequently fed into another convolutional block to produce keypoint reliability scores \( \mathbf{R}_A \in \mathbb{R}^{H/8 \times W/8 \times 1}\). These scores are combined with the keypoint heatmap to detect keypoints.

For the keypoint detection, to preserve spatial resolution without compromising speed, the input image is represented as a 2D grid where each grid cell consists of an \( 8 \times 8 \) pixel patch, which is reshaped into a 64-dimensional feature vector. After four fast \( 1 \times 1 \) convolutional layers, a keypoint embedding \( \mathbf{K}_A \in \mathbb{R}^{H/8 \times W/8 \times (64+1)} \) is obtained, where each unit \( \mathbf{k}_{i,j} \in \mathbf{K}_A \) encodes the keypoint distribution logits. Keypoints are then classified into one of the 64 possible positions within 
\(
\mathbf{k}_{i,j} \in \mathbb{R}^{65}
\)
with an additional bin to account for the absence of keypoints. During inference, this additional bin is discarded, and the heatmap is reinterpreted as an \( 8 \times 8 \) cell representation to get \( \tilde{\mathbf{K}}_A \in \mathbb{R}^{H \times W \times 1} \).

To ensure accurate keypoint detection, the keypoint reliability map \( \mathbf{R}_A \) is interpolated to match the input image size, yielding \( \tilde{\mathbf{R}}_A \in \mathbb{R}^{H \times W \times 1} \). A threshold \( \sigma \) is set for keypoint extraction: for each point in the region, if the condition \( \widetilde{k}_{i,j} \cdot \widetilde{r}_{i,j} > \sigma \) is met, the point is identified as a keypoint, and its corresponding descriptor \( d_{i,j} \) is retrieved from \( \mathbf{D}_A \).  

The same procedure is applied for another image region \( B \), yielding keypoints and descriptors for both areas. Keypoint matching is performed using a mutual nearest neighbor approach, followed by the RANSAC algorithm to estimate the homography transformation matrix \( \mathbf{H} \). Based on \( \mathbf{H} \), the UAV image center point \( \mathbf{C}_U \) is mapped to its corresponding point in the satellite image, thus calculating the position of UAV.

This lightweight architecture ensures high-speed feature extraction while being computationally efficient. Although the exclusive use of stacked CNNs may limit the ability to capture high-level semantic information, this issue is mitigated as the feature extraction is performed only in small localized regions, given that the coarse matching stage has already established correspondences between image regions.
\subsection{Absolute Visual Localization Pipeline}
We revisit the problem of cross-source image matching for UAV visual localization and propose a hierarchical framework that operates in a coarse-to-fine manner. Specifically, we first leverage high-level semantic features to rapidly identify and align corresponding regions between UAV and satellite images, which are robust to fine-grained appearance variations yet rich in contextual information. Then, we extract fine-grained geometric features within these localized regions to achieve precise pixel-level matching.
Furthermore, we incorporate an image retrieval module as the first stage, which retrieves geographically relevant satellite images from a large-scale database based on the UAV's captured image. By combining retrieval and matching, we establish a complete localization pipeline.

In summary, our proposed pipeline integrates three key components: an image retrieval module, a coarse matching module guided by deep semantic cues and structural constraints, and a fine-grained matching module based on a lightweight convolutional network. These components collectively form a complete solution for UAV absolute visual localization. The effectiveness of the proposed framework is demonstrated through experiments on both public and self-collected datasets, as detailed in the next section.
\begin{table*}[htbp]
    \centering
    \caption{Localization performance of various algorithms on the AerialVL dataset. The AerialVL\_Comb.1 and AerialVL\_Comb.2 results are directly taken from \cite{he2024aerialvl}, as their code is not publicly available.}
    \label{tab:AV}
    \begin{tabular}{@{}c|c|ccccc|cccccc|c@{}}
        \toprule
        \multicolumn{2}{c|}{} & \multicolumn{5}{c|}{\textbf{Short-Term Trajectories}} & \multicolumn{6}{c|}{\textbf{Long-Term Trajectories}} & \\
        \cmidrule(lr){3-7} \cmidrule(lr){8-13}
         \textbf{Methods}& \textbf{Metrics} & ST.1 & ST.2 & ST.3 & ST.4 & ST.5 & LT.1 & LT.2 & LT.3 & LT.4 & LT.5 & LT.6 & \textbf{Avg.} \\
        \midrule
             \multirow{2}{*}{SIFT\cite{lowe2004sift}} 
               & Succ.Rate& 0.33 & 0.39 & 0.34 & 0.42 & 0.44 & 0.39 & 0.45 & 0.39 & 0.45 & 0.47 & 0.40 & 0.41 \\
             
              & MLE (m)& 31.27 & 29.54 & 29.53 & 27.22 & 27.09 & 28.85 & 25.91 & 28.67 & 26.97 & 25.93 & 28.02 & 28.09 \\
             \multirow{2}{*}{SuperPoint\cite{detone2018superpoint}} 
                & Succ.Rate& 0.40 & 0.45 & 0.43 & 0.48 & 0.56 & 0.43 & 0.55 & 0.38 & 0.53 & 0.56 & 0.46 & 0.48 \\
             
                & MLE (m)& 29.92 & 27.71 & 27.38 & 26.37 & 24.86 & 27.19 & 23.69 & 28.95 & 24.18 & 23.60 & 26.73 & 26.42 \\
             \multirow{2}{*}{Deep-LK\cite{goforth2019gps}} 
                & Succ.Rate& \textbf{0.53} & 0.48 & 0.51 & 0.56 & 0.44 & 0.43 & 0.48 & 0.29 & 0.54 & 0.45 & 0.57 & 0.48 \\
             
                & MLE (m)& 27.21 & 26.94 & 25.91 & 25.25 & 27.32 & 26.85 & 26.63 & 31.86 & 24.63 & 26.92 & 26.94 & 26.95 \\
        \midrule
             \multirow{2}{*}{AerialVL\_Comb.1 \cite{he2024aerialvl}} 
                & Succ.Rate& - & - & - & - & 0.71 & - & - & - & - & 0.41 & 0.70 & -\\
             
               & MLE (m)& - & - & - & - & 20.01 & - & - & - & - & 26.92 & 17.89 & -\\
             \multirow{2}{*}{AerialVL\_Comb.2 \cite{he2024aerialvl}} 
                & Succ.Rate& - & - & - & - & 0.80 & - & - & - & - & 0.83 & \textbf{0.88} & -\\
             
                & MLE (m)& - & - & - & - & 22.27 & - & - & - & - & 14.22 & 17.50 & -\\
             \multirow{2}{*}{\textbf{Ours}} 
                & Succ.Rate& \textbf{0.53} & \textbf{0.80} & \textbf{0.93} & \textbf{0.91} & \textbf{0.92} & \textbf{0.70} & \textbf{0.96} & \textbf{0.63} & \textbf{0.79} & \textbf{0.92} & 0.85 & \textbf{0.81} \\
             
                & MLE (m)& \textbf{27.10} & \textbf{23.03} & \textbf{15.82} & \textbf{18.69} & \textbf{19.62} & \textbf{18.84} & \textbf{14.75} & \textbf{21.76} & \textbf{17.21} & \textbf{14.05} & \textbf{15.20} & \textbf{18.73} \\
        \bottomrule
    \end{tabular}
\end{table*}
\begin{table*}[htbp]
    \centering
    \caption{Localization performance evaluation of various algorithms on the CS-UAV dataset}
    \label{tab:CS}
    \begin{tabular}{@{}c|c|ccccccccc|c@{}}
        \toprule
        \textbf{Method} & \textbf{Metrics} & T.1 & T.2 & T.3 & T.4 & T.5 & T.6 & T.7 & T.8 & T.9 & Avg. \\ 
        \midrule
            \multirow{2}{*}{SIFT\cite{lowe2004sift}} 
            & Succ.Rate & 0.64 & 0.63 & 0.43 & 0.51 & 0.46 & 0.65 & 0.60 & 0.55 & 0.54 & 0.56 \\
            & MLE (m)   & 22.16 & 21.62 & 27.07 & 25.42 & 27.69 & 22.01 & 23.57 & 23.51 & 24.72 & 24.21\\
            \multirow{2}{*}{SuperPoint\cite{detone2018superpoint}} 
            &  Succ.Rate &0.64 & 0.64 & 0.43 & 0.51 & 0.46 & 0.58 & 0.58 & 0.54 & 0.54 & 0.54 \\
            &  MLE (m)   &22.19 & 21.40 & 27.07 & 25.34 & 27.71 & 23.28 & 24.44 & 23.98 & 24.78 & 24.64\\
            \multirow{2}{*}{Deep-LK\cite{goforth2019gps}}
            &  Succ.Rate &0.39 & 0.43 & 0.40 & 0.42 & 0.28 & 0.41 & 0.32 & 0.31 & 0.31 & 0.36 \\
            &  MLE (m)   &27.77 & 27.69 & 27.97 & 28.08 & 32.50 & 27.53 & 30.08 & 29.86 & 30.56 & 29.78 \\
        \midrule
            \multirow{2}{*}{\textbf{Ours}} 
            &  Succ.Rate &\textbf{0.97} & \textbf{0.95} & \textbf{0.93} & \textbf{0.79} & \textbf{0.87} & \textbf{0.98} & \textbf{0.96} & \textbf{0.97} & \textbf{0.87} & \textbf{0.93} \\
            &  MLE (m)  &\textbf{14.69} & \textbf{15.75} & \textbf{19.30} & \textbf{20.71} & \textbf{20.57} & \textbf{16.55} & \textbf{14.50} & \textbf{14.19} & \textbf{20.15} & \textbf{17.71}\\
        \bottomrule
    \end{tabular}
\end{table*}
\section{Experiments}
To validate the effectiveness of our proposed visual localization pipeline, we conduct extensive experiments using both public and self-collected UAV datasets. This section provides details about the datasets, training procedures, and evaluation protocols. All experiments are executed on a workstation equipped with an Intel Core i9-14900K CPU and an NVIDIA RTX A6000 GPU.
\subsection{Dataset}
We utilize two public datasets and one self-collected dataset to train and evaluate our methods, including \textbf{UAV-Visloc} \cite{xu2024uavvisloc}, \textbf{AerialVL} \cite{he2024aerialvl} and the newly introduced \textbf{CS-UAV}.

\textbf{UAV-Visloc} is employed for training. It contains 6,742 UAV images captured at varying altitudes and orientations across 11 geographically diverse locations in China, using both fixed-wing and multi-terrain drones. Each UAV image is geotagged with precise metadata (latitude, longitude, altitude, timestamp), accompanied by a corresponding satellite map for each region.

\begin{figure}[t]
      \centering
      \includegraphics[width=\linewidth]{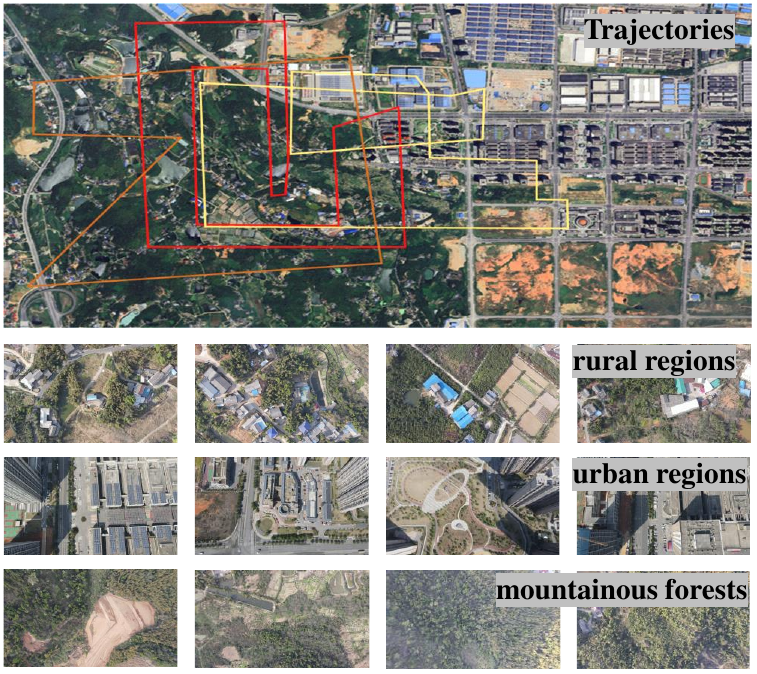}
      \caption{The top image illustrates several example UAV flight trajectories collected in Changsha, China. These trajectories (marked in different colors) span diverse terrains, including rural regions, urban regions, and mountainous forests. The lower section provides representative image samples captured from the UAV during flights, clearly depicting the variety of environmental conditions encountered. High-precision onboard GPS records provide accurate ground-truth localization for each frame in the dataset.}
      \label{CS-UAV}
   \end{figure}

\textbf{AerialVL} and \textbf{CS-UAV} are used for evaluation. AerialVL is a UAV-based visual localization benchmark collected in Qingdao, China. It comprises 11 flight sequences ranging from 3.7 km to 11 km, covering diverse terrain types, altitudes, and lighting conditions. The dataset includes corresponding satellite images, facilitating fine-grained UAV-satellite alignment.
The CS-UAV dataset was collected in Changsha, China, using a DJI Mavic 2 drone. It comprises 9 flight trajectories averaging approximately 10 kilometers each, covering diverse environments such as urban areas, rural regions, and mountainous forests. High-precision GPS data was simultaneously recorded onboard, providing accurate ground-truth positioning information. This dataset is publicly accessible at [link]. Detailed information about the dataset and illustrative examples are shown in Fig.\ref{CS-UAV}.
\subsection{Model Training}
During training, the coarse matching network and fine matching network are trained separately, which is adopted for two reasons: first, due to the non-differentiable nature of the matching process, designing an end-to-end training strategy is challenging; second, since the two networks are specifically designed to focus on different levels of features, training them jointly in an end-to-end manner could hinder convergence and make the training process unstable.

For the coarse matching network, positive samples are constructed by pairing each UAV image in UAV-Visloc with a satellite patch extracted based on its GPS location. Negative samples are randomly generated by pairing UAV images with non-corresponding satellite regions. These sample pairs are used to train the semantic-aware and structure-constrained matching module, as described in Section~\ref{Method}.

For the fine matching network, we generate a synthetic dataset by applying random geometric and photometric transformations (e.g., rotation, scaling, color jitter) to UAV images of UAV-Visloc. Since transformation parameters are known, pixel-level correspondences can be directly used for supervised learning.

\subsection{Localization Evaluation}
We evaluate UAV localization performance using two metrics proposed in \cite{he2024aerialvl}: Mean Localization Error (MLE) and Success Rate. Localization is considered successful when the error is below 25 meters, and any error exceeding 50 meters is regarded as drift. Accordingly, we compute the success rate and mean localization error within the non-drift segments of the flight trajectory.

\textbf{Overall Comparison.} To evaluate the overall effectiveness of our proposed framework, we compare it against two baseline methods reported in \cite{he2024aerialvl} across three representative flight trajectories of AerialVL. We do not include additional trajectories in the comparison, as the source code of this work is not publicly available, making it difficult to reproduce their results. As summarized in Table~\ref{tab:AV}, our method consistently achieves superior localization performance, even though both baseline approaches incorporate relative localization techniques in addition to absolute localization.

\textbf{Module-Level Comparison.} To further assess the effectiveness of our hierarchical matching framework, we conduct comparative experiments with several representative matching methods (SIFT\cite{lowe2004sift}, SuperPoint \cite{detone2018superpoint}, Deep-LK\cite{goforth2019gps}) used by prior UAV visual localization studies \cite{tingshua_uav,goforth2019gps,he2024aerialvl}. All methods are integrated into our pipeline, using the same image retrieval component with only the matching module being varied. For a fair comparison, all models are trained on UAV-specific datasets, except for SIFT, which is used directly. Experimental results on the AerialVL and CS-UAV datasets are presented in Tables~\ref{tab:AV} and \ref{tab:CS}, respectively.

\textbf{Ablation studies.} To further assess the contribution of the refinement of image matching, we introduce a simplified variant, where the final position is derived solely from the top-1 retrieved satellite image without matching. Moreover, we evaluate an ablation version of our method that excludes the semantic-aware matching module to quantify its contribution to the overall performance. As shown in Table~\ref{tab:ablation_combined}, these variants suffer from higher localization errors and lower success rates, particularly in challenging scenarios.

\textbf{Analysis.} Our complete framework achieves superior performance across most trajectories, demonstrating significant improvements in both success rate and localization accuracy. Compared with baseline methods such as SuperPoint and Deep-LK, our approach exhibits stronger robustness to drastic appearance changes, primarily due to its hierarchical matching strategy that effectively integrates both coarse and fine alignment stages. The retrieval-only variant, which relies solely on the initial image retrieval without further refinement, performs poorly in terms of both the success rate and mean localization error, as it only provides a rough localization estimate without precise geometric alignment. Furthermore, removing the Semantic-Aware and Structure-Constrained Region Matching Module (SASCM) results in a noticeable drop in performance, underscoring its critical role in enhancing matching accuracy by incorporating high-level semantic cues and geometric consistency. These results collectively validate the effectiveness of our proposed design components in addressing the challenges of visual localization under GNSS-denied conditions.
\begin{table}[htbp]
    \centering
    \caption{Ablation study results on both AerialVL and CS-UAV datasets. All values represent averages computed over all flight trajectories in each dataset.}
    \label{tab:ablation_combined}
    \begin{tabular}{@{}c|cc|cc@{}}
        \toprule
        \multirow{2}{*}{\textbf{Methods}} & \multicolumn{2}{c|}{\textbf{AerialVL}} & \multicolumn{2}{c}{\textbf{CS-UAV}} \\
        & \textbf{Succ.Rate} & \textbf{MLE (m)} & \textbf{Succ.Rate} & \textbf{MLE (m)} \\
        \midrule
        \textbf{Retrieval-Only} & 0.41 & 28.11 & 0.54 & 24.20 \\
        \textbf{Ours w/o SASCM} & 0.68 & 22.14 & 0.72 & 21.53 \\
        \textbf{Ours} & \textbf{0.81} & \textbf{18.73} & \textbf{0.93} & \textbf{17.71} \\
        \bottomrule
    \end{tabular}
\end{table}
\section{CONCLUSIONS}
In this paper, we introduce a hierarchical cross-source image matching method designed for UAV absolute localization, which integrates a semantic-aware and structure-constrained coarse matching module with a lightweight fine-grained matching module. Building upon the matching method, an image retrieval module is employed as the first stage, constructing a robust and complete pipeline for UAV absolute localization without reliance on relative localization techniques. Experimental evaluations on public benchmark datasets and a newly introduced CS-UAV dataset demonstrate superior accuracy and robustness of the proposed method under various challenging conditions, confirming its effectiveness.

While the proposed method achieves notable improvements in localization accuracy, it inevitably introduces some computational burden. Future work will investigate methods such as neural network acceleration and early-exit mechanisms to alleviate this and enhance runtime efficiency without compromising accuracy.





\end{document}